\documentclass[sigconf]{acmart}

\renewcommand\footnotetextcopyrightpermission[1]{}
\usepackage{booktabs}
\usepackage{amsmath}
\usepackage{amsfonts}
\usepackage{graphicx}
\usepackage{multirow}
\usepackage{balance}

\title{CLAIM: Leading Open-domain Active Clarification of Large Language Models with Uncertainty Measurement}

\author{Kuangzhao Yang}
\affiliation{
  \institution{Gaoling School of Artificial Intelligence, Renmin University of China}
  \city{Beijing}
  \country{China}
}
\email{yangkuangzhao050519@ruc.edu.cn}

\author{Ziliang Zhao}
\affiliation{
  \institution{Gaoling School of Artificial Intelligence, Renmin University of China}
  \city{Beijing}
  \country{China}
}
\email{zhaoziliang@ruc.edu.cn}

\author{Zhicheng Dou}
\authornote{Zhicheng Dou is the corresponding author.}
\affiliation{
  \institution{Gaoling School of Artificial Intelligence, Renmin University of China}
  \city{Beijing}
  \country{China}
}
\email{dou@ruc.edu.cn}

\begin{document}

\begin{abstract}
In open-domain human–computer interaction scenarios, large language models (LLMs) frequently encounter user queries that are ambiguous or incomplete.
In such cases, directly producing an answer often leads to overgeneralized, erroneous, or low-information responses.
In contrast, asking clarifying questions can substantially improve interaction quality.
However, existing approaches still rely heavily on manually annotated data or preference alignment to address two fundamental challenges: when clarification is necessary, and which aspect of the query should be clarified.
This reliance incurs high annotation costs and limits generalization.
To address these challenges, we propose CLAIM, an uncertainty-driven framework for active clarification learning in open-domain settings.
CLAIM eliminates the need for explicit human preference annotations by quantifying query uncertainty through the entropy induced by answer disagreements across multiple models.
This uncertainty signal is then used to construct high-quality synthetic data, enabling the training of a unified clarification decision model through a combination of supervised learning and reinforcement learning.
Specifically, we propose an entropy-driven synthetic data generation pipeline that integrates entropy-based uncertainty estimation with semantic clustering and reasoning-based judgments, enabling reliable automatic annotation of clarification requirements.
To train CLAIM, we formulate the clarification process as a structured decision generation problem and adopt a training paradigm that combines supervised fine-tuning (SFT) with group-relative policy optimization (GRPO).
Experimental results demonstrate that CLAIM can learn stable and generalizable clarification strategies without relying on manually labeled data, offering a low-cost and robust solution for proactive understanding in real-world open-domain interactions with LLMs.
\end{abstract}

\begin{CCSXML}
<ccs2012>
   <concept>
       <concept_id>10002951.10003317.10003338.10003341</concept_id>
       <concept_desc>Information systems~Language models</concept_desc>
       <concept_significance>500</concept_significance>
       </concept>
 </ccs2012>
\end{CCSXML}

\ccsdesc[500]{Information systems~Language models}

\keywords{Open-domain Clarification; Large Language Models; Uncertainty Estimation; Active Decision Learning; Synthetic Data
}

\maketitle

\section{Introduction}
\begin{figure}[t]
    \centering
    \includegraphics[width=\linewidth]{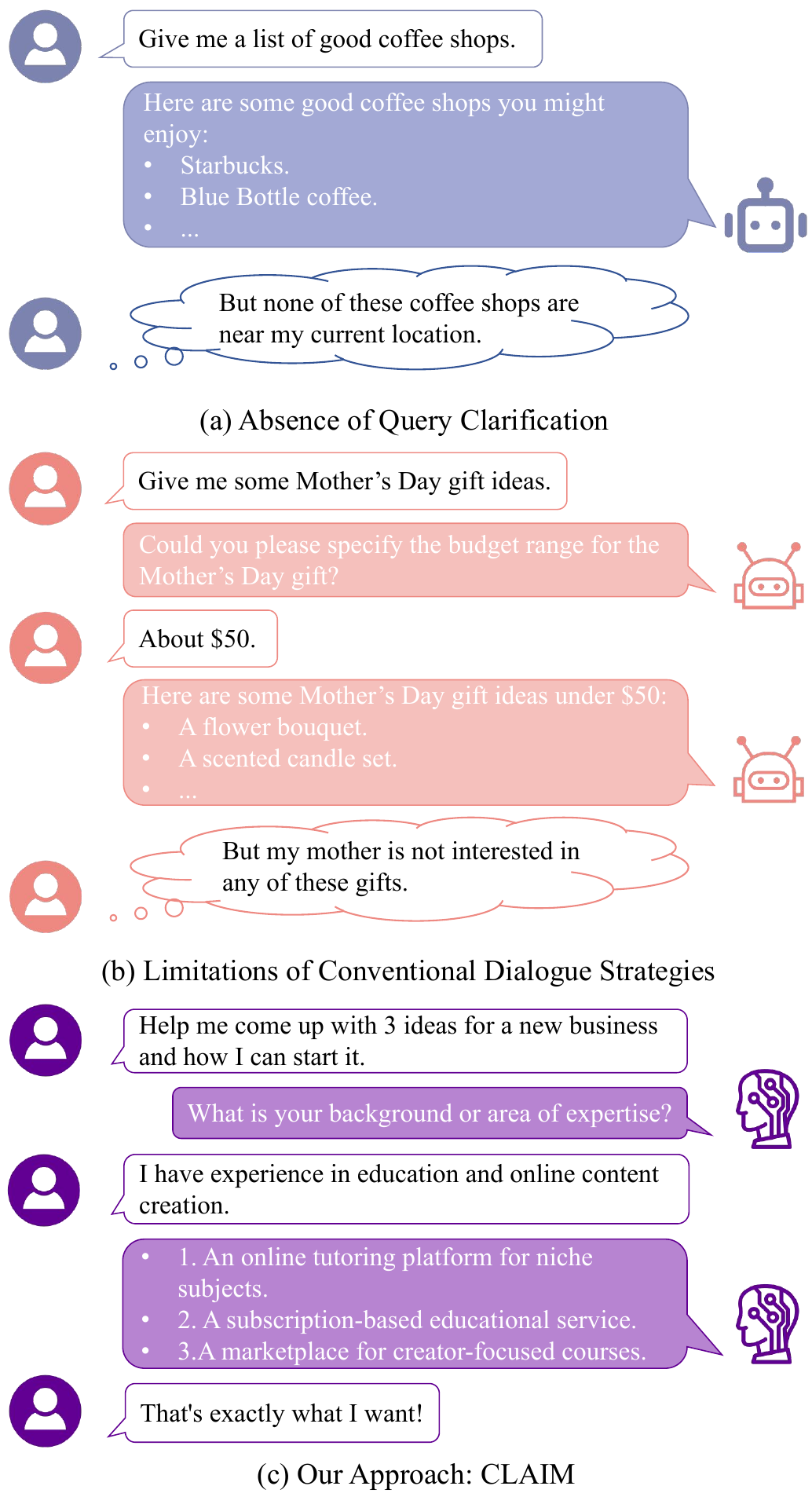}
    \caption{Examples of clarification in human--LLM interaction.
    (a) Missing clarification leads to generic answers.
    (b) Clarification on uninformative dimensions fails.
    (c) CLAIM aligns clarification with user intent and improves outcomes.}
    \label{fig:intro_case}
\end{figure}

In real-world human--LLM interaction scenarios, user queries are often not formulated as fully specified and precisely defined requests~\cite{zhang2024clamber,qian2024implicit}.
Instead, users tend to express their needs using short, high-level, or ambiguous natural language, such as ``Give me some gift ideas,'' or ``Recommend some coffee shops.''
Such queries frequently omit key constraints, preferences, or contextual details that are essential for producing accurate and user-aligned responses. 
When a large language model responds directly to these underspecified queries without access to critical missing information, the resulting answers are often generic, overly broad, or misaligned with the user’s true intent, as illustrated in Figure~\ref{fig:intro_case}(a).
In contrast, proactively asking clarifying questions and guiding users to provide additional information allows large language models to progressively narrow the intent space and resolve underlying ambiguity. 
Through this interactive process, models can better identify user goals, filter irrelevant interpretations, and generate responses that are both more accurate and more informative. 
Prior studies have shown that such active clarification substantially improves the accuracy and relevance of subsequent responses, particularly in open-domain settings where user intent is highly variable~\cite{lee-etal-2023-asking,zhang2024futureturns}.
As a result, active clarification has been widely recognized as an important mechanism for improving the quality of open-domain human--LLM interactions~\cite{ma2025ambigchat}.

Despite the importance of active clarification, enabling reliable clarification in open-domain human--LLM interaction remains highly challenging.
First, a model must determine whether a user query is sufficiently clear to support a direct response.
Not all queries benefit from clarification, and unnecessary follow-up questions can interrupt the interaction flow and degrade user experience~\cite{aliannejadi2019asking,aliannejadi2020convai3clariq,zhao2025clarilm}.
Second, even when a query is genuinely ambiguous or underspecified, the model must choose which missing information dimension to clarify~\cite{lee-etal-2023-asking}.
A single query may simultaneously lack information about location, user preferences, or specific constraints, yet these dimensions often differ substantially in their importance and information gain.
Clarifying along a less informative dimension while overlooking more decisive factors frequently leads to unsatisfactory outcomes, as shown in Figure~\ref{fig:intro_case}(b).
Consequently, prioritizing clarification questions that target dimensions with higher information gain is widely regarded as a key factor in determining the effectiveness of clarification~\cite{yuan2024multimodal,zhao2024intentaware,ma2025ambigchat}.

Existing approaches typically rely on manually annotated data, handcrafted rules, or preference-based supervision to train clarification decision models~\cite{aliannejadi2019asking,aliannejadi2020convai3clariq,zhao2025clarilm}.
However, in open-domain settings, the diversity and long-tail nature of user queries make such supervision costly to obtain and difficult to scale~\cite{majumder2021global}.
Moreover, clarification decisions are often implicitly entangled with language generation, leading to blurred decision boundaries and limited controllability~\cite{sekulic2021facetdriven,lee-etal-2023-asking}.
These limitations restrict the generalization and applicability of existing clarification methods in real-world interaction scenarios.

We revisit open-domain clarification from a different perspective, motivated by a simple yet widely observed phenomenon.
\textbf{When a user query is well-specified, different large language models tend to produce semantically consistent responses; in contrast, when the query is ambiguous or underspecified, the responses generated by different models often diverge substantially at the semantic level}.
Such semantic divergence reflects uncertainty in the model’s understanding of the query and provides a strong intrinsic signal indicating whether clarification is needed.
This observation is consistent with prior findings showing that semantic disagreement among multiple sampled generations correlates strongly with model uncertainty and answer unreliability~\cite{wang2023selfconsistency,kadavath2022language}. It is also aligned with recent policy-discriminative learning work such as POLAR, which uses trajectories from diverse policies to expose behavior-level differences that are difficult to obtain from a single policy alone~\cite{dou2025pretrained}.
By leveraging this signal, a model can ask targeted clarification questions that effectively narrow the intent space and lead to satisfactory outcomes, as illustrated in Figure~\ref{fig:intro_case}(c).

Building on this intuition, we propose \textbf{CLAIM} (Leading Open-domain Active Clarification of Large Language Models with Uncertainty Measurement), an open-domain active clarification framework driven by model-intrinsic uncertainty.
CLAIM prompts multiple heterogeneous large language models to generate candidate answers for the same user query and performs semantic clustering over these responses, characterizing query uncertainty through the entropy of the resulting answer distribution~\cite{shannon1948entropy}.
Based on this uncertainty signal, CLAIM automatically determines whether a query is worth clarifying and constructs training instances that capture uncertainty changes before and after clarification.
This process relies solely on model-generated outputs and requires no human annotations or external preference signals, enabling clarification supervision to be obtained in a scalable and low-cost manner.

At the training level, CLAIM formulates open-domain clarification as a conditional generation policy learning problem~\cite{zhao2025clarilm,ouyang2022training}.
The model is initialized via supervised fine-tuning (SFT) to acquire stable clarification decision and generation behaviors, and is further refined with group relative policy optimization (GRPO) to improve decision consistency and robustness near the clarification boundary~\cite{rafailov2023dpo,shao2024deepseekmath,kuhn2023semanticuncertainty}.
When clarification is required, the model explicitly identifies the semantic dimension to be clarified and generates a targeted clarification question grounded in that dimension~\cite{sekulic2021facetdriven,majumder2021global}.

In summary, our contributions can be summarized as follows:

(1) We propose \textbf{CLAIM}, an \textbf{uncertainty-driven open-domain active clarification framework} that enables large language models to adapt their interaction strategies based on intrinsic uncertainty in understanding user queries;

(2) We design an \textbf{entropy-driven synthetic data generation pipeline} that constructs high-quality training data \emph{without any human annotations}, while explicitly modeling uncertainty changes induced by clarification;

(3) We formulate clarification as an explicit decision learning problem and train the model with a combination of \textbf{SFT} and \textbf{GRPO}, improving the controllability and generalization of clarification decisions while maintaining overall system simplicity.

\section{Related Work}

\subsection{Clarification for LLMs}
In open-domain human--LLM interactions, clarification mitigates brittle guessing under underspecification.
ClariLM~\cite{zhao2025clarilm} synthesizes open-domain clarification data and trains models for when-to-ask and what-to-ask via supervised learning and preference optimization, while outcome-aware training optimizes for future-turn success and encourages clarification when it improves answerability under multiple interpretations~\cite{zhang2024futureturns}.
In tool-augmented settings, Ask-when-Needed~\cite{wang2025askwhenneeded} asks for missing/unclear arguments before tool use to improve robustness to noisy instructions.

Several benchmarks have examined clarification behavior in LLMs and consistently reveal a gap between answering performance and effective information acquisition.
CLAMBER~\cite{zhang2024clamber} and ClarQ-LLM~\cite{gan2024clarqllm} evaluate models’ ability to recognize ambiguity and obtain missing information through clarification across open-domain and task-oriented settings, showing that strong response generation alone does not translate into reliable clarification.
AR-Bench~\cite{zhou2025arbench} and QuestBench~\cite{li2025questbench} further demonstrate that even when models can solve fully specified tasks, they often fail to identify what information is missing or which question should be asked under incomplete inputs.
Beyond ambiguity resolution, clarification has also been explored for preference and constraint elicitation, such as sequential funnel-style questioning~\cite{montazeralghaem2025preference} and ambiguity reduction in code generation~\cite{wu2023codeclarify}.
Taken together, these studies suggest that the core challenge in clarification lies not only in generating follow-up questions, but in reliably deciding \emph{when} clarification is needed and \emph{which information dimension} to target—highlighting the need for principled signals, such as uncertainty, to guide clarification decisions in open-domain human--LLM interaction.

\subsection{Clarification in Other Domains}
Clarification has been studied as information acquisition across conversational systems.
In conversational IR, question selection before re-ranking~\cite{aliannejadi2019asking} and large-scale resources with engagement signals (MIMICS)~\cite{zamani2020mimics} support clarification research; later work also generates questions from weak supervision such as query reformulations under supervised and reinforcement learning.
In multi-turn search, Qulac~\cite{aliannejadi2019qulac} enables facet-specific evaluation, and utility-driven ranking connects question choice to expected information gain, formalized by neural EVPI~\cite{rao2018evpi}.
Related formulations appear in exploratory conversational search~\cite{liu2024mining} and active learning/adaptive information acquisition~\cite{rothe2017question_as_program}.
Beyond retrieval, clarification supports interactive recommendation and decision support via preference elicitation~\cite{zhang2022multiinterest}, resolves ambiguity/missing constraints in multimodal or embodied settings~\cite{white2021open}, and is central to conversational machine reading (ShARC)~\cite{saeidi2018sharc} and ambiguity-aware QA (AmbigQA)~\cite{min2020ambigqa}.

\subsection{Entropy and Uncertainty-Based Methods}
Uncertainty-driven clarification requires quantifying ambiguity to decide whether and how to ask; information-theoretic utility and EVPI-based ranking are classic formulations~\cite{rao2018evpi}.
For LLMs, uncertainty is often measured in the \emph{answer space} via sampling and disagreement, where surface diversity may miss semantic conflict; semantic uncertainty addresses this by clustering generations by meaning and computing entropy over clusters (semantic entropy)~\cite{kuhn2023semanticuncertainty,farquhar2024semanticentropy}. Beyond repeated decoding from a single model, recent policy-discriminative learning also shows the value of sampling outputs from diverse policies to capture cross-policy behavioral differences~\cite{dou2025pretrained}.
To reduce sampling cost, SEPs approximate semantic entropy from internal representations~\cite{kossen2024seps}, while Cleanse uses embedding-space clustering as a proxy for semantic consistency~\cite{joo2025cleanse}; together they motivate practical, model-agnostic alternatives to probability-based confidence~\cite{kadavath2022language}.
For clarification-specific modeling, CLARINET distills an information-gain objective conditioned on a retrieval distribution~\cite{chi2024clarinet}.
These results motivate CLAIM: leverage \textbf{multi-model disagreement}, aggregate via \textbf{semantic clustering}, and compute \textbf{entropy over clusters} for clarification decisions and data construction.

\section{CLAIM}
\begin{figure*}[t]
    \centering
    \includegraphics[width=0.8\textwidth]{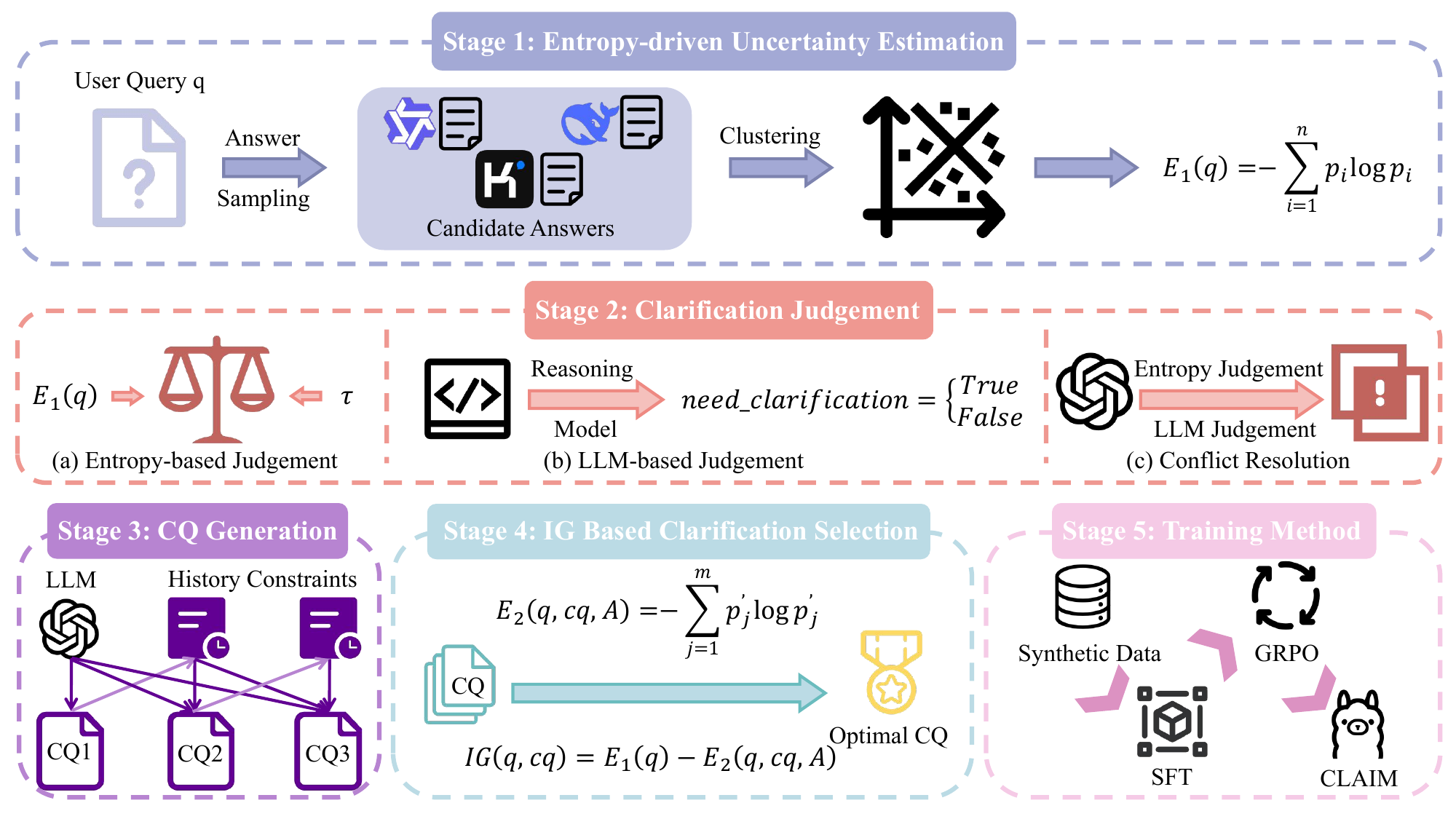}
    \caption{The overall framework of CLAIM, consisting of entropy-driven uncertainty estimation, clarification judgement, clarifying question generation, information gain-based selection, and SFT/GRPO-based training.}
    \label{fig:framework}
\end{figure*}
CLAIM addresses clarification decision-making in open-domain scenarios through uncertainty-driven synthetic data construction. 
To operationalize the proposed framework, we implement CLAIM as an agent-style offline pipeline, referred to as CLAIM-Agent. 
It is important to distinguish the two: \textbf{CLAIM-Agent} is a system-level data-construction agent that executes uncertainty estimation, clarification judgement, conflict arbitration, and clarifying question selection, whereas \textbf{CLAIM} is the trained single-model policy used for online inference. 
Thus, the multi-model and multi-call cost of CLAIM-Agent is paid only during offline synthetic data construction; after SFT/GRPO training, CLAIM performs one standard model inference per user query. 
This section formalizes the task and describes the synthetic data construction process and training methodology.
The corresponding implementation code and prompts are released in the repository\footnote{\url{https://github.com/ykun49365/CLAIM-final}}. 

\subsection{Problem Formulation}
We consider the clarification decision-making task in single-turn open-domain human--LLM interactions. 
Given a user query $q$, which may take the form of a question, instruction, or information request, the input can be ambiguous or lack critical information. 
The model is required to make a decision based on the current input, determining whether to provide a direct answer or to initiate a clarification.

Formally, the model output $r$ belongs to one of two categories: a direct answer $a$ or a clarifying question $c$. The clarification decision can thus be formulated as:
\begin{equation}
\text{CLAIM}(q) = r \in \{a, c\},
\end{equation}
where choosing $a$ indicates that the model considers the information in $q$ sufficient, while choosing $c$ indicates that additional interaction is needed to obtain missing information.

When the model decides to ask for clarification, it must further determine the clarification dimension, that is, which type of missing information to query. 
This decision directly affects the efficiency of subsequent interactions and the quality of information acquisition. 
Therefore, clarification decision-making involves not only deciding whether to clarify, but also selecting the appropriate clarification dimension.
This formulation aligns with recent agent-based frameworks that treat interaction as a sequence of decision-making steps combining reasoning and action.

\subsection{Entropy-driven Uncertainty Estimation}
In clarification decision-making, a core challenge is to determine whether a user query contains sufficient information for a stable and consistent answer without relying on strong assumptions. 
To this end, CLAIM models question uncertainty from the perspective of the answer space, by analyzing the distribution of answers generated by different models for the same query to capture potential ambiguity.

Specifically, given a user query $q$, we use $k_1$ different LLMs to independently generate a set of candidate direct answers:
\begin{equation}
\mathcal{A}(q) = \{a_1, a_2, \ldots, a_{k_1}\}.    
\end{equation}
We then project all candidate answers into a shared semantic representation space and perform clustering based on semantic similarity, resulting in $n$ clusters of semantically consistent answers. 
Let $c_i$ denote the number of answers in the $i$-th cluster. 
The corresponding cluster probability is defined as:
\begin{equation}
p_i = \frac{c_i}{k_1}, \quad \sum_{i=1}^{n} c_i = k_1.    
\end{equation}
Based on this distribution, we quantify the uncertainty of the user query $q$ using entropy:
\begin{equation}
E_1(q) = - \sum_{i=1}^{n} p_i \log p_i .    
\end{equation}
Intuitively, when different models generate semantically consistent answers, the clustering is concentrated and the entropy is low, indicating that the query is well-specified. 
In contrast, when answers are divided into multiple semantically distinct clusters and the distribution becomes more dispersed, the entropy increases, suggesting that the query may be ambiguous or lack critical information. 
This entropy signal serves as a continuous measure of uncertainty and provides an important basis for subsequent clarification decisions and clarification strategy selection.

\subsection{Clarification Judgement}

\subsubsection{Entropy-based Judgement}
Based on the answer distribution entropy $E_1(q)$ obtained in the previous section, CLAIM first derives a threshold-based clarification judgement that converts continuous uncertainty into an initial decision.
Specifically, we introduce a threshold $\tau$ and define:
\begin{equation}
y_{\text{ent}}(q) = \mathbb{I}\big(E_1(q) \ge \tau\big),
\end{equation}
where $y_{\text{ent}}(q)=1$ indicates that clarification is needed, and $y_{\text{ent}}(q)=0$ otherwise.
A larger $E_1(q)$ reflects stronger semantic divergence among model-generated answers, suggesting higher uncertainty in query interpretation, while lower entropy typically corresponds to more consistent and well-specified queries. 
We use a single global threshold $\tau=0.45$ for all datasets and discuss its rationale in Appendix~\ref{AP:threshold}.

\subsubsection{LLM-based Judgement}
While entropy captures answer-level disagreement, it does not directly assess whether a query is semantically complete.
To complement this signal, CLAIM employs a reasoning model that evaluates the query itself and judges whether critical information required for a precise answer is missing.
This judgement focuses on semantic completeness (e.g., ambiguous references or missing constraints) and produces an independent clarification signal that complements entropy-based uncertainty.

\subsubsection{Conflict Resolution}
In practice, entropy-based and LLM-based judgements may disagree.
Some queries exhibit high answer entropy yet remain answerable under reasonable assumptions, whereas others yield consistent answers while still lacking essential information.
Relying on either signal alone is therefore insufficient.

When such disagreement occurs, CLAIM invokes an additional judgement model to arbitrate the conflict.
This model takes as input both the uncertainty characteristics reflected by the answer distribution and the semantic analysis underlying the LLM-based judgement, and produces the final clarification decision.
By explicitly resolving judgement conflicts through a dedicated arbitration step, CLAIM mitigates failure modes near the clarification boundary and yields more robust clarification judgements, which are then used for downstream question generation and data construction. 
Appendix~\ref{AP:conflict} reports the proportion of samples requiring this arbitration step.

\subsection{Clarifying Question Generation}
For queries requiring clarification, CLAIM generates clarifying questions to acquire missing information.
Rather than exhaustively enumerating all possibilities, it produces a small set of diverse candidates that differ in the information they aim to elicit.

Concretely, given a user query $q$, CLAIM repeatedly invokes the same generation model to produce multiple clarifying questions.
Each generated question is accompanied by a dimension label that characterizes the primary aspect of missing information it targets.
This label is not drawn from a predefined dimension taxonomy, but serves as an auxiliary annotation produced during generation.

To encourage diversity across candidates, CLAIM applies history-based constraints in later generation steps by conditioning on previously generated questions and their dimension labels, preventing the reuse of already covered clarification dimensions.
As a result, the generated clarifying questions target complementary aspects of the query and form the candidate set for subsequent selection based on uncertainty reduction.

\subsection{Information Gain Based Clarification Selection}
After generating multiple candidate clarifying questions, CLAIM evaluates the effectiveness of different clarification dimensions in reducing query uncertainty. 
To this end, we quantify the value of a clarification dimension by comparing the change in answer distribution uncertainty before and after clarification.

Concretely, given a user query $q$ and a candidate clarifying question $cq$, together with a simulated user answer $A$, we again employ the same $k_1$ large language models used in the initial stage to generate a set of post-clarification candidate answers. 
These answers are then semantically clustered, and the entropy of the post-clarification answer distribution is computed in the same manner as in the initial uncertainty modeling:
\begin{equation}
E_2(q, cq, A) = - \sum_{j=1}^{m} p'_j \log p'_j ,
\end{equation}
where $m$ denotes the number of clusters obtained after clarification, and $p'_j$ represents the probability of the $j$-th cluster.

Based on this, we define the reduction in uncertainty introduced by the clarifying question $cq$ as the information gain($IG$):
\begin{equation}
IG(q, cq) = E_1(q) - E_2(q, cq, A).    
\end{equation}
The information gain captures the effectiveness of a clarifying question in resolving ambiguity or supplementing missing critical information.

For multiple candidate clarifying questions generated for the same user query, CLAIM selects the one with the highest information gain as the optimal clarifying question. 
Through this information gain–based selection mechanism, CLAIM prioritizes clarifying questions that most effectively reduce uncertainty, thereby constructing high-quality clarification decision data.

\subsection{Training Method}
Based on the uncertainty-driven synthetic data constructed in the previous stages, we adopt a two-stage training paradigm to learn a stable and consistent clarification decision policy. 
The purpose of training is to internalize clarification behaviors into a single model, so that clarification decisions can be made directly at inference time without relying on repeated multi-model interactions or agent-style pipelines. 
In the first stage, SFT is used to enable the model to acquire basic clarification behaviors from automatically constructed supervision signals. 
In the second stage, GRPO is applied to further sharpen the decision boundary for queries with high semantic uncertainty. 
This training design fully exploits synthetic data generated in the previous steps and enables effective modeling of clarification strategies without relying on human preference annotations.

\subsubsection{SFT}
In the supervised fine-tuning stage, the model is trained as a conditional generation policy given a user query.
The model outputs a structured representation that specifies the clarification decision together with the associated generation content.
When clarification is required, the generation proceeds by first identifying a single semantic dimension and then producing a clarification question grounded in that dimension; otherwise, the model directly produces a final answer.

The supervision signal is derived from target output sequences constructed during the synthetic data generation process. 
These sequences encode both the clarification decision and its associated generation content, where the selection of clarification questions is designed to enhance the discriminability of answers after clarification. 
By modeling the above behaviors as a unified sequence generation task, the model can learn a complete clarification strategy within a conditional generation framework.

Formally, let the training dataset be defined as
\begin{equation}
\mathcal{D} = \{(x_i, y_i)\}_{i=1}^{N},    
\end{equation}
where \(x_i\) denotes a user query and \(y_i\) denotes the corresponding target output sequence. 
The model is parameterized as a conditional probability distribution \(p_\theta(y \mid x)\), and the objective of supervised fine-tuning is to minimize the autoregressive negative log-likelihood loss:
\begin{equation}
\mathcal{L}_{\text{SFT}}(\theta)
= - \mathbb{E}_{(x,y)\sim \mathcal{D}}
\left[
\sum_{t=1}^{|y|} \log p_\theta(y_t \mid x, y_{<t})
\right].    
\end{equation}
By optimizing this objective, the model learns the clarification triggering patterns implicit in the synthetic data as well as the corresponding language generation behaviors, providing a stable initialization for the subsequent policy optimization stage.

\subsubsection{GRPO}
Although SFT effectively conveys global clarification decision signals, the model may still exhibit instability or bias when query semantics lie close to the clarification boundary. 
To further improve decision consistency in regions of high uncertainty, we introduce GRPO in the second stage to refine the learned policy.

In this stage, the model is treated as a stochastic policy $\pi_\theta$. 
For each input query $x$, the current policy generates a group of candidate outputs under a fixed sampling configuration:
\begin{equation}
\mathcal{Y}(x) = \{y^{(1)}, y^{(2)}, \dots, y^{(K)}\}.    
\end{equation}
Each candidate output is evaluated by a deterministic reward function $r(x, y)$, which measures the extent to which the generated result aligns with the target decision induced by the synthetic data process in terms of clarification decision consistency, semantic plausibility, and structural validity.

To avoid dependence on the absolute scale of rewards, the group-wise average reward is used as a baseline:
\begin{equation}
\bar{r}(x) = \frac{1}{K} \sum_{k=1}^{K} r(x, y^{(k)}).    
\end{equation}
Based on this, the relative advantage of each output is defined as:
\begin{equation}
A(x, y^{(k)}) = r(x, y^{(k)}) - \bar{r}(x).    
\end{equation}
During policy updates, we adopt a clipped objective based on probability ratios to ensure training stability. 
Let $\pi_{\theta_{\text{old}}}$ denote the policy from the previous iteration, and the probability ratio is defined as:
\begin{equation}
\rho_\theta(x, y) = \frac{\pi_\theta(y \mid x)}{\pi_{\theta_{\text{old}}}(y \mid x)}.    
\end{equation}
The objective of group relative policy optimization is then given as:
\begin{equation}
\begin{aligned}
\mathcal{L}_{\text{GRPO}}(\theta)
= - \mathbb{E}_{x}\Bigg[
\frac{1}{K}\sum_{k=1}^{K}
\min\Big(
\rho_\theta(x, y^{(k)})\, A(x, y^{(k)}), \\
\qquad
\text{clip}\!\left(
\rho_\theta(x, y^{(k)}),\, 1-\epsilon,\, 1+\epsilon
\right)\, A(x, y^{(k)})
\Big)
\Bigg].
\end{aligned}
\end{equation}
where $\epsilon$ is a clipping coefficient that constrains the magnitude of policy updates. It effectively prevents excessive updates under high-variance reward signals and improves training stability.

In practice, the policy optimization stage primarily focuses on queries with high semantic uncertainty, which typically correspond to cases with highly dispersed candidate answers or conflicting decisions in the synthetic data process. 
By imposing group-relative constraints on these critical samples, the model gradually learns a clearer and more robust clarification decision boundary.

\section{Experiments}
Additional implementation details and training configurations are deferred to Appendix~\ref{AP3}.

\begin{table*}[t]
\centering
\caption{Main evaluation results of CLAIM (8B) and baseline models on three test sets. The best result for each metric is marked in \textbf{bold} and the second best result for each metric is \underline{underlined}.}
\label{tab:main_results}
\resizebox{\textwidth}{!}{%
\begin{tabular}{llcccccccccccc}
\toprule
 &  & \multicolumn{4}{c}{\textbf{ClariLM-test}} & \multicolumn{4}{c}{\textbf{IN3}} & \multicolumn{4}{c}{\textbf{CLAMBER}} \\
\cmidrule(lr){3-6}\cmidrule(lr){7-10}\cmidrule(lr){11-14}
Group & Model
& \multicolumn{2}{c}{Clari.\ Necessity} & \multicolumn{2}{c}{Clari.\ Quality}
& \multicolumn{2}{c}{Clari.\ Necessity} & \multicolumn{2}{c}{Clari.\ Quality}
& \multicolumn{2}{c}{Clari.\ Necessity} & \multicolumn{2}{c}{Clari.\ Quality} \\
\cmidrule(lr){3-4}\cmidrule(lr){5-6}\cmidrule(lr){7-8}\cmidrule(lr){9-10}\cmidrule(lr){11-12}\cmidrule(lr){13-14}
 &  & Acc & F1 & CDA & CQSS & Acc & F1 & CDA & CQSS & Acc & F1 & CDA & CQSS \\
\midrule

\multirow{5}{*}{LLM}
& Llama-3.1-8B & 61.35 & 68.07 & 25.04 & 50.31 & 78.70 & 87.83 & 30.52 & 56.90 & 54.87 & 55.55 & 32.98 & 63.82 \\
& Qwen3-8B & 55.72 & 66.89 & 21.24 & 49.12 & 82.86 & 89.89 & 31.58 & 57.95 & 53.14 & 58.18 & 28.48 & 58.87 \\
& Qwen3-14B & 73.76 & 77.79 & 26.33 & 53.17 & 79.63 & 87.36 & 18.95 & 53.78 & 62.25 & 63.74 & 44.03 & 59.67 \\
& Qwen3-32B & 76.90 & 82.86 & 54.44 & \textbf{70.20} & 84.26 & 90.71 & 60.00 & 73.83 & 57.80 & 62.16 & 56.40 & 66.49 \\
& DeepSeek-V3 & 77.83 & 80.27 & 47.01 & 61.42 & 75.93 & 84.34 & 53.68 & 70.05 & 58.96 & 54.68 & 52.78 & 65.34 \\
\midrule

\multirow{2}{*}{LRM}
& QwQ-32B & 77.36 & 81.04 & 55.01 & 66.62 & 77.78 & 86.21 & 49.47 & 69.95 & 57.94 & 58.59 & 60.21 & 65.54 \\
& DeepSeek-R1 & 53.82 & 65.46 & 53.55 & 66.02 & 73.08 & 83.97 & 61.05 & 70.13 & 58.16 & 62.98 & 61.34 & \textbf{67.12} \\
\midrule

\multirow{5}{*}{SFT}
& SFT-Entropy only & 74.60 & 79.15 & 53.88 & 62.67 & 79.63 & 88.17 & 64.21 & 72.34 & 48.59 & 49.23 & 47.03 & 60.93 \\
& SFT-LLM only & 73.40 & 77.96 & 53.31 & 56.39 & 76.85 & 86.77 & 61.05 & 70.76 & 51.41 & 51.80 & 50.59 & 62.14 \\
& SFT-without IG & 77.65 & 81.69 & 43.05 & 52.17 & 80.56 & 88.89 & 50.52 & 67.64 & 60.40 & 61.06 & 44.97 & 59.84 \\
& SFT-IN3~\cite{qian2024implicit} & 74.35 & 78.44 & 49.03 & 59.33 & \textbf{89.81} & \underline{94.12} & \underline{70.53} & \textbf{75.78} & 56.00 & 56.63 & 55.90 & 61.85 \\
& SFT-Full & 79.60 & 82.86 & \underline{55.17} & 67.38 & 85.19 & 91.40 & \textbf{71.58} & \underline{74.65} & 61.99 & 62.40 & 62.27 & \underline{66.91} \\
\midrule

Related Work & ClariLM~\cite{zhao2025clarilm} & \underline{81.25} & \underline{85.48} & 52.50 & 63.55 & \underline{89.72} & \textbf{94.36} & 66.32 & 72.68 & \underline{64.23} & \underline{67.61} & \underline{62.89} & 65.36 \\
\midrule

\multirow{2}{*}{Our}
& CLAIM-Agent & 71.95 & \textbf{91.58} & 51.94 & 63.67 & 85.19 & 91.58 & 68.42 & 71.39 & 59.62 & 66.63 & 58.15 & 60.43 \\
& CLAIM & \textbf{81.85} & 84.97 & \textbf{56.79} & \underline{69.54} & 87.04 & 92.55 & 63.16 & 72.23 & \textbf{65.18} & \textbf{68.10} & \textbf{63.71} & 65.47 \\

\bottomrule
\end{tabular}%
}
\end{table*}

\subsection{Datasets}
We evaluate the model’s open-domain clarification ability on three representative datasets, all of which are used exclusively for evaluation. 
First, we adopt the synthetic clarification dataset constructed in ClariLM denoted as ClariLM-test, and only use its test set as an evaluation benchmark~\cite{zhao2025clarilm}.
This dataset is automatically generated and organized around latent missing information facets in user queries, with structured instances that explicitly distinguish between clarification and non-clarification cases, making it suitable for evaluating whether a model makes appropriate clarification decisions under controlled conditions. 
Second, we use the IN3 (Intention-in-Interaction) dataset as an evaluation benchmark for task-oriented interactive scenarios~\cite{qian2024implicit}. 
IN3 is grounded in real-world vague user instructions and provides systematic annotations on task ambiguity, missing critical details, and their relative importance, enabling the evaluation of clarification behavior in interactive task contexts. 
Although IN3 provides training data for model development, we do not use its training data and evaluate exclusively on its test set. 
Finally, we include CLAMBER as a general open-domain clarification benchmark~\cite{zhang2024clamber}. 
CLAMBER covers a broad range of open-domain topics and focuses on evaluating a model’s ability to identify uncertainty in natural language queries and to generate high-quality clarification questions. 
Together, these three complementary datasets allow us to comprehensively evaluate clarification performance across synthetic data, task-oriented interaction settings, and general open-domain scenarios.

\subsection{Evaluation Metrics}
We evaluate the model’s clarification ability from two complementary aspects: clarification necessity and clarifying question quality. 
For clarification necessity, we formulate the task as a binary classification problem, where the model determines whether a given user query requires clarification. 
We report Accuracy (ACC) and F1-score (F1) to measure overall decision correctness and performance under class imbalance.
For samples where the model decides that clarification is necessary, we further assess the quality of the generated clarifying questions. 
Specifically, clarifying question quality is evaluated from two complementary perspectives. 
First, we define Clarification Dimension Accuracy (CDA) to measure whether the generated clarifying question focuses on the correct clarification dimension. 
We employ an independent large language model as an evaluator to judge whether the clarification dimension of the generated question matches the ground-truth dimension. 
A match is assigned a score of 1, and a mismatch a score of 0. 
The final CDA score is computed as the average over all samples:
\begin{equation}
\mathrm{CDA} = \frac{1}{N}\sum_{i=1}^{N} \mathbb{I}(d_i^{\text{pred}} = d_i^{\text{gt}}),
\end{equation}
where $d_i^{\text{pred}}$ and $d_i^{\text{gt}}$ denote the predicted and ground-truth clarification dimensions for the $i$-th sample, respectively. 
Second, we measure Clarifying Question Semantic Similarity (CQSS) to evaluate the semantic closeness between the generated clarifying question and its ground-truth counterpart. 
Specifically, we compute the cosine similarity between their vector representations for each sample and report the average similarity across all samples:
\begin{equation}
\mathrm{CQSS} = \frac{1}{N}\sum_{i=1}^{N} 
\cos\bigl( \mathbf{e}(q_i^{\text{pred}}), \mathbf{e}(q_i^{\text{gt}}) \bigr),
\end{equation}
where $\mathbf{e}(\cdot)$ denotes Qwen3-Embedding-8B, and cosine similarity is computed after normalizing the two question embeddings. 
Together, these metrics enable a systematic evaluation of the model’s clarification ability in terms of both deciding whether clarification is needed and generating appropriate clarifying questions.
The evaluation prompts are released with the code repository.

\subsection{Baseline Models}
We select a diverse set of baselines to comprehensively evaluate the effectiveness of CLAIM for open-domain clarification. 
These baselines span different model scales (8B, 32B, etc.), architectural paradigms (direct generation vs. explicit reasoning), and training strategies (zero-shot, supervised fine-tuning, and agent-based execution), enabling systematic comparison across both clarification decision and question generation behaviors.
This design allows us to analyze clarification performance under diverse modeling assumptions while maintaining a unified evaluation protocol.
%Overall, we will systematically analyze clarification performance across model scale (8B, 32B, etc.), architectural design (direct generation vs. explicit reasoning), and training strategy (zero-shot, supervised fine-tuning, and agent-based execution), establishing rigorous and interpretable baselines for evaluating open-domain clarification capability.

We first include a group of general-purpose LLMs as direct-answering baselines, including Llama-3.1-8B, Qwen3-8B, Qwen3-14B, Qwen3-32B, and DeepSeek-V3. 
These models are evaluated in a zero-shot setting without any explicit clarification decision mechanism, serving to characterize the default behavior of mainstream LLMs when handling ambiguous or underspecified user queries.

Next, to examine whether strong reasoning ability alone is sufficient for reliable clarification, we consider two reasoning-oriented language reasoning models (LRMs), namely QwQ-32B and DeepSeek-R1. 
Although these models are not specifically trained for clarification, their chain-of-thought reasoning capabilities allow them to explicitly analyze semantic completeness and missing information, providing a useful comparison for understanding the role of reasoning in clarification decision-making.

We further report a group of supervised fine-tuned clarification models (SFT-based models) as training-based baselines. 
These models share the same base architecture but differ in the construction of supervision signals during the SFT stage, allowing controlled ablation of key components in the CLAIM framework.
Specifically, \textbf{SFT-Entropy only} is trained using clarification decisions derived solely from entropy-based uncertainty estimation, without incorporating LLM-based semantic judgement. 
Conversely, \textbf{SFT-LLM only} relies exclusively on LLM-based judgement to determine whether clarification is required, without using entropy signals.
To evaluate the effect of clarification question selection, we include \textbf{SFT-without IG}, which follows the full clarification judgement procedure but selects clarifying questions without information gain (IG)–based ranking.
In addition, \textbf{SFT-IN3} is fine-tuned using the training set provided by the IN3 dataset and focuses on clarification in task-oriented interaction scenarios. 
In contrast, \textbf{SFT-Full} is trained on the complete CLAIM synthetic dataset constructed with the full uncertainty-driven pipeline, covering both clarification-required and non-clarification scenarios in open-domain settings.

We also include \textbf{ClariLM} as a representative prior open-domain clarification method that is trained with large-scale supervised and preference-based data to jointly model clarification decisions and question generation~\cite{zhao2025clarilm}. 
We reproduce ClariLM following the original paper and evaluate it under the same evaluation protocol and metrics for fair comparison.
Unless otherwise specified, none of the SFT-based models incorporate GRPO, ensuring that comparisons at this stage reflect the effects of different supervision signals and SFT strategies rather than reinforcement-based policy optimization.
We do not include CLARINET or EVPI-based rankers as direct baselines because their inputs and objectives differ from CLAIM: they assume candidate clarifying questions or retrieval distributions and optimize question selection/ranking, whereas CLAIM jointly decides whether clarification is needed and what open-domain question to ask without a predefined candidate set.

\textbf{CLAIM-Agent} is additionally included as a special baseline to validate the soundness of the proposed uncertainty-driven synthetic data construction and clarification workflow. CLAIM-Agent directly executes the full clarification pipeline in an agent-style manner without supervised fine-tuning or policy optimization. 
Its performance therefore reflects the effectiveness of the proposed clarification process itself, rather than improvements induced by learned model parameters.

\subsection{Overall Results}

Table~\ref{tab:main_results} reports the main evaluation results of CLAIM and all baseline models on three benchmarks.
Overall, CLAIM achieves \textbf{state-of-the-art (SOTA) or near-SOTA} performance on the majority of metrics across all datasets, demonstrating the effectiveness of its uncertainty-driven framework in jointly deciding \emph{when} clarification is needed and \emph{what} information should be requested.

Comparing different model categories, zero-shot LLMs and reasoning oriented LRMs exhibit large performance variance across datasets.
Although some large-scale models achieve competitive results on individual metrics, their overall performance remains \textbf{unstable}, suggesting that general language generation ability or reasoning capability alone is insufficient for reliable clarification.
In contrast, SFT-based models are more \textbf{robust}, and CLAIM maintains \textbf{strong and balanced} performance across all three benchmarks.

Compared with the prior open-domain clarification method ClariLM, \textbf{CLAIM matches or outperforms it on most metrics}, particularly on CLAMBER and ClariLM-test.
Notably, this performance is achieved using only approximately \textbf{10k} uncertainty-constructed training instances, whereas ClariLM relies on around \textbf{120k} supervised and preference-annotated examples~\cite{zhao2025clarilm}.
This comparison highlights the \textbf{data efficiency and scalability of CLAIM} in open-domain clarification settings.

\subsection{Further Analysis}
\paragraph{Effect of Clarification Judgement Signals.}
Among SFT-based models, SFT-Entropy only and SFT-LLM only rely on a \textbf{single} clarification judgement signal during supervision.
While both variants outperform zero-shot baselines, their performance remains consistently lower than that of SFT-Full across datasets.
For example, on ClariLM-test, SFT-Full achieves an accuracy of 79.60 and a CDA score of 55.17, compared to 74.60/53.88 for SFT-Entropy only and 73.40/53.31 for SFT-LLM only.
This gap indicates that entropy-based uncertainty estimation and LLM-based semantic judgement capture complementary aspects of query ambiguity, and that relying on either signal alone is insufficient for robust clarification decisions.

\paragraph{Effect of Information Gain for Clarifying Question Selection.}
Comparing SFT-without IG with SFT-Full isolates the effect of information gain–based clarification question selection.
Although both models use the same clarification judgement mechanism, SFT-without IG shows clear degradation in clarification quality metrics, especially \textbf{CDA and CQSS}.
On ClariLM-test, removing information gain reduces CDA from 55.17 to 43.05 and CQSS from 67.38 to 52.17, despite similar necessity prediction performance.
This result suggests that explicitly selecting clarifying questions based on uncertainty reduction plays a critical role in identifying more informative clarification dimensions.

\paragraph{Effect of GRPO.}
Finally, comparing SFT-Full and CLAIM isolates the contribution of group relative policy optimization (GRPO).
While SFT-Full already exhibits strong overall performance, CLAIM further improves decision stability and clarification quality on multiple benchmarks.
For instance, CLAIM improves ClariLM-test accuracy from 79.60 to 81.85 and CDA from 55.17 to 56.79, and also increases CLAMBER accuracy from 61.99 to 65.18.
This improvement suggests that GRPO effectively refines the clarification decision boundary, particularly for queries with high semantic uncertainty.

\paragraph{CLAIM-Agent Analysis.}
CLAIM-Agent, which directly executes the full clarification \textbf{pipeline} without model training, achieves competitive performance across several metrics.
On IN3, CLAIM-Agent achieves an F1 score of 91.58, and on CLAMBER it reaches an accuracy of 59.62, demonstrating its competitiveness with trained SFT models.
This result validates the soundness of the proposed uncertainty-driven data construction and clarification workflow.
Although CLAIM-Agent underperforms CLAIM in some metrics, CLAIM benefits from GRPO-based training with group-relative advantages, enabling the model to internalize clarification strategies through self-sampling and relative comparison during training.

\paragraph{Generalization Across Domains.}
The comparison between SFT-IN3 and SFT-Full further illustrates differences in generalization behavior.
SFT-IN3 achieves strong performance on the IN3 benchmark, with an accuracy of 89.81, but degrades noticeably on ClariLM-test and CLAMBER, where its accuracy drops to 74.35 and 56.00, respectively.
In contrast, SFT-Full maintains consistently competitive results across all benchmarks, achieving accuracies of 79.60 on ClariLM-test, 85.19 on IN3, and 61.99 on CLAMBER.
These results indicate that uncertainty-driven synthetic data construction enables robust clarification learning without overfitting to a specific domain. The comparison also reduces the attribution concern that improvements come merely from task-specific post-training: SFT-IN3, SFT-Entropy only, SFT-LLM only, SFT-without IG, and SFT-Full use comparable training recipes but differ in supervision source and selection mechanism.

\subsection{LLM-as-a-Judge and Human Evaluation}
\begin{figure}[t]
\centering
\includegraphics[width=\linewidth]{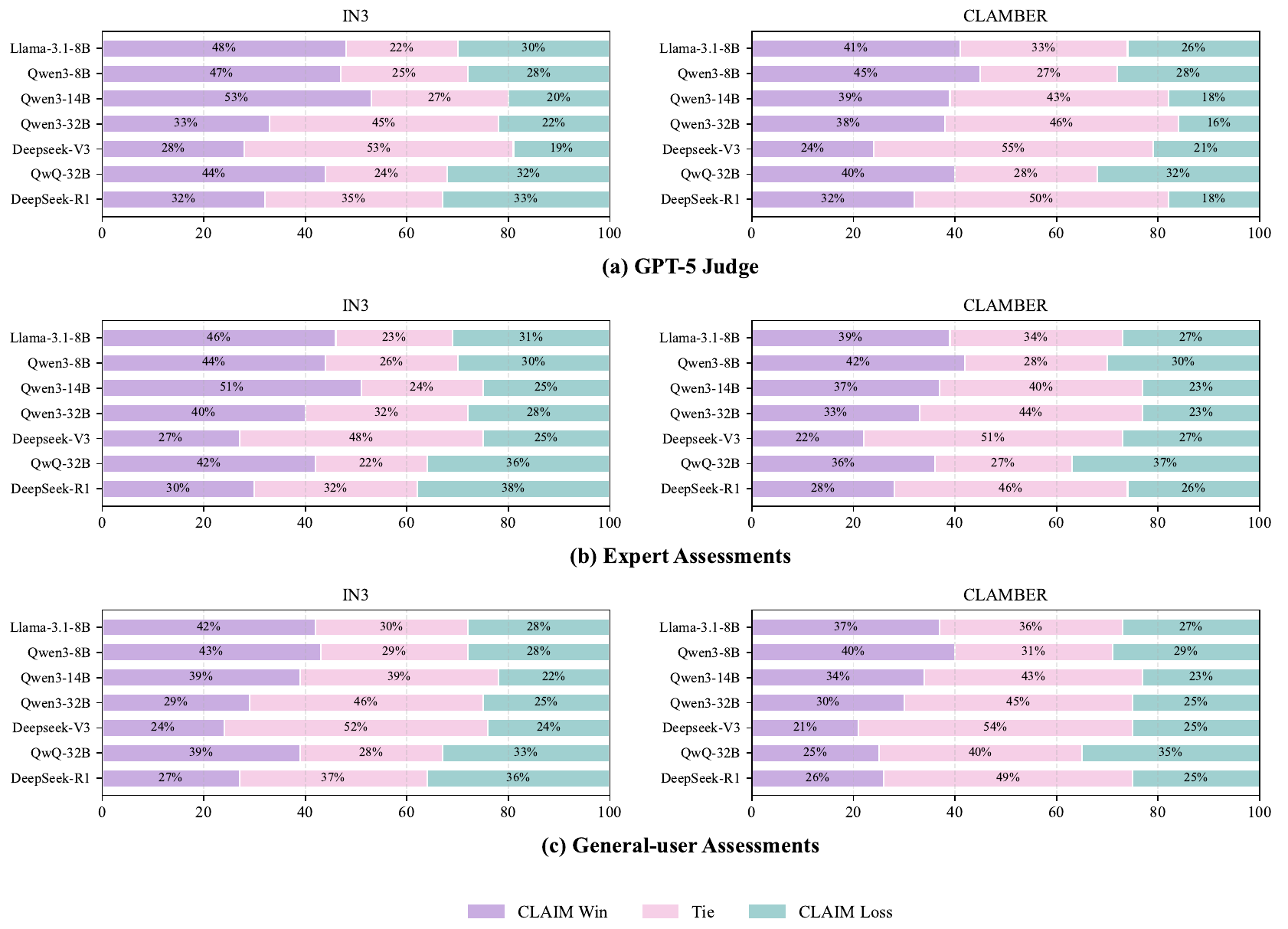}
\caption{Pairwise comparative evaluation of CLAIM against representative baselines under GPT-5 judge, expert, and general-user assessments. Each panel reports Win/Tie/Loss percentages over 100 sampled instances.}
\label{fig:judge_human_eval}
\end{figure}

While automated metrics provide a quantitative assessment of clarification performance, they are limited in capturing overall usefulness and interaction quality.
Following prior work on LLM-based comparative evaluation~\cite{gu2024survey_llm_as_judge}, we conduct pairwise LLM-as-a-Judge evaluation using GPT-5, where each judge input contains the user query and two anonymized model outputs, and the judge returns Win, Tie, or Lose for CLAIM against a baseline.
To calibrate this automatic evaluation, we further conduct a human study under the same pairwise protocol.
Three expert annotators and a group of general users each evaluate 100 randomly sampled instances per baseline on IN3 and CLAMBER.
Figure~\ref{fig:judge_human_eval} summarizes the GPT-5 judge and human evaluation results together.

The human results are consistent with the GPT-5 evaluation trend: CLAIM obtains more wins than losses against most baselines, especially smaller zero-shot LLMs and reasoning models.
The advantage is smaller against stronger 32B/V3 models, where ties increase, indicating that human judges often view both outputs as similarly useful when the baseline already asks a reasonable clarification question.
Overall, the agreement between LLM-as-a-Judge and human evaluation supports the claim that CLAIM improves user-perceived clarification behavior rather than only optimizing automatic metrics.

\section{Conclusion}
In this paper, we propose CLAIM, an uncertainty-driven framework for open-domain clarification in large language models. 
By leveraging semantic disagreement among multiple heterogeneous LLMs, CLAIM estimates answer uncertainty and synthesizes large-scale clarification data without relying on domain-specific resources. 
The uncertainty-aware clarification selection mechanism enables the model to effectively determine when clarification is necessary and which question to ask. 
We further adopt a two-stage training paradigm combining SFT and GRPO to enhance decision stability under high uncertainty. 
Extensive experiments on multiple benchmarks demonstrate that CLAIM consistently outperforms strong baseline models in both clarification necessity detection and clarifying question quality. 
This paper focuses on single-turn clarification; extending CLAIM to multi-turn interaction requires explicit dialogue-state tracking, history-dependent uncertainty estimation, and planning over future turns, which we leave for future work.
These results highlight the effectiveness of uncertainty-aware modeling for general-domain clarification and suggest promising directions for building more adaptive and reliable human--LLM interaction systems.

\appendix

\section{Additional Analysis and Implementation Details} \label{AP3}
The uncertainty-driven data construction and clarification selection processes follow the methodology described in Section~3.
To estimate answer distribution uncertainty, CLAIM samples candidate responses from multiple heterogeneous large language models ($k_1 = 5$), including \textbf{DeepSeek-V3}, \textbf{Qwen3-32B}, \textbf{GLM-4-32B-0414}, \textbf{Kimi-K2-Instruct-0905}, and \textbf{Ling-flash-2.0}.
Semantic clustering and entropy computation are performed using representations from \textbf{Qwen3-Embedding-8B} with cosine similarity.
For clarifying question generation and user response simulation, CLAIM uses \textbf{DeepSeek-V3} to generate $k_2 = 3$ candidate clarifying questions per query.
All generation processes adopt a unified decoding temperature of 0.7.
The prompt templates used by CLAIM-Agent and the evaluation scripts are released in the code repository.

\paragraph{Entropy Threshold.} \label{AP:threshold}
In all experiments, we set $\tau=0.45$ as a single global threshold without tuning it separately for different benchmarks.
This value follows a simple entropy-based intuition under our default $k_1=5$ setting: when only one model produces a semantically different answer, the smallest non-unanimous cluster distribution is $(0.8,0.2)$, whose entropy is $-0.8\log 0.8 - 0.2\log 0.2 \approx 0.50$.
We therefore set $\tau$ slightly below this value so that one-model disagreement is treated as a weak but actionable uncertainty signal.
The final decision is not determined by entropy alone, since LLM-based judgement and conflict resolution further correct boundary cases.

\paragraph{LLM Calls and Token Cost.}
CLAIM-Agent is used only for offline data construction.
For a non-clarification query, the pipeline uses at most $k_1$ direct-answer calls, one LLM-based judgement call, and one optional arbitration call, i.e., no more than 7 LLM calls under $k_1=5$.
For a clarification query, the pipeline additionally uses $k_2$ calls that jointly generate clarifying questions and simulated user answers, plus $k_1k_2$ post-clarification answer calls, resulting in no more than 25 LLM calls under $k_1=5,k_2=3$.
In our implementation, constructing 1k synthetic examples consumes approximately 5.7M tokens.
These calls are embarrassingly parallel and are not required at deployment time, where CLAIM runs as a single model.

\paragraph{Training Details.}
For training, we adopt \textbf{Meta-Llama-3.1-8B-Instruct} as the base model. 
The supervised fine-tuning (SFT) stage is implemented using \textbf{LLaMA-Factory} with \textbf{LoRA} parameter-efficient fine-tuning, where the LoRA rank is set to 8 and the LoRA alpha to 16. 
We use a learning rate of $5\times10^{-5}$ and train for 3 epochs.
The maximum sequence length is set to 2048, training uses bf16 precision and Flash Attention, the per-device batch size is 2 with 8 gradient accumulation steps, and gradient norms are clipped to 1.0.
In the second stage, we apply GRPO using the \textbf{TRL} framework, initialized from the SFT-trained model.
For each query, the policy samples $K=4$ candidate outputs under temperature 0.7.
Policy updates use clipping coefficient $\epsilon=0.2$ and KL coefficient $\beta=0.01$ with AdamW learning rate $1\times10^{-6}$.
All local training experiments are conducted on \textbf{four NVIDIA RTX 6000 Ada Generation GPUs}.

\section{Clarification-Ratio Diagnostics} \label{AP:ratio}
We additionally report diagnostic clarification ratios for different uncertainty-construction strategies.
The ground-truth column denotes the clarification-required ratio in each benchmark, Single-model multi-sample denotes repeated sampling from one model, and Multi-model single-sample denotes one sample from each heterogeneous model.
This diagnostic is intended to test whether repeated sampling from a single strong model can replace heterogeneous model disagreement.
Our preliminary observation is that within-model sampling may remain semantically concentrated for queries with strong priors, even under higher temperature, while heterogeneous models more often expose distinct interpretations. 
Semantic clustering is then applied before entropy computation to reduce surface-form sampling noise.

\begin{table}[h]
\centering
\caption{Clarification-required ratios under different uncertainty sampling strategies.}
\label{tab:clarification_ratio}
\begin{tabular}{lccc}
\toprule
Dataset & Ground Truth & Single-model & Multi-model \\
\midrule
ClariLM-test & 61.90 & 53.85 & 77.80 \\
IN3 & 87.96 & 34.26 & 73.15 \\
CLAMBER & 50.00 & 31.01 & 61.09 \\
\bottomrule
\end{tabular}
\end{table}

\section{Conflict Arbitration Statistics} \label{AP:conflict}
Table~\ref{tab:conflict_ratio} reports the proportion of samples for which entropy-based judgement and LLM-based judgement disagree and therefore require arbitration.
The non-trivial ratios across both training and evaluation data support the need for conflict resolution rather than relying on a single judgement signal.

\begin{table}[h]
\centering
\caption{Samples requiring conflict arbitration between entropy-based and LLM-based judgements.}
\label{tab:conflict_ratio}
\begin{tabular}{lcc}
\toprule
Dataset & Quantity & Percentage \\
\midrule
Training & 5242 & 52.42 \\
CLAMBER & 1190 & 37.16 \\
ClariLM-test & 932 & 46.60 \\
IN3 & 30 & 27.78 \\
\bottomrule
\end{tabular}
\end{table}

\section{GenAI Usage Disclosure}
In this paper, GenAI is primarily used for synthesizing partial data in the methodology, with the relevant details explicitly stated in the main text. 
Additionally, while GenAI is not employed in drafting the manuscript from scratch, it (GPT-5.2) is utilized for error checking (including grammar, tense, etc.) after manual completion.

\bibliographystyle{ACM-Reference-Format}
\balance
\bibliography{references}

@misc{gan2024clarqllm,
  title        = {ClarQ-LLM: A Benchmark for Models Clarifying and Requesting Information in Task-Oriented Dialog},
  author       = {Gan, Yujian and Li, Changling and Xie, Jinxia and Wen, Luou and Purver, Matthew and Poesio, Massimo},
  year         = {2024},
  eprint       = {2409.06097},
  archivePrefix= {arXiv},
  primaryClass = {cs.CL},
  url          = {https://arxiv.org/abs/2409.06097},
  note         = {arXiv preprint}
}

@inproceedings{wang2025askwhenneeded,
  title     = {Learning to Ask: When LLM Agents Meet Unclear Instruction},
  author    = {Wang, Wenxuan and Shi, Juluan and Ling, Zixuan and Chan, Yuk-Kit and Wang, Chaozheng and Lee, Cheryl and Yuan, Youliang and Huang, Jen-tse and Jiao, Wenxiang and Lyu, Michael R.},
  booktitle = {Proceedings of the 2025 Conference on Empirical Methods in Natural Language Processing},
  year      = {2025},
  pages     = {21773--21784},
  address   = {Suzhou, China},
  publisher = {Association for Computational Linguistics},
  doi       = {10.18653/v1/2025.emnlp-main.1104},
  url       = {https://aclanthology.org/2025.emnlp-main.1104/}
}

@misc{zhou2025arbench,
  title        = {From Passive to Active Reasoning: Can Large Language Models Ask the Right Questions under Incomplete Information?},
  author       = {Zhou, Zhanke and Feng, Xiao and Zhu, Zhaocheng and Yao, Jiangchao and Koyejo, Sanmi and Han, Bo},
  year         = {2025},
  eprint       = {2506.08295},
  archivePrefix= {arXiv},
  primaryClass = {cs.CL},
  url          = {https://arxiv.org/abs/2506.08295},
  note         = {arXiv preprint}
}

@misc{li2025questbench,
  title        = {QuestBench: Can LLMs ask the right question to acquire information in reasoning tasks?},
  author       = {Li, Belinda Z. and Kim, Been and Wang, Zi},
  year         = {2025},
  eprint       = {2503.22674},
  archivePrefix= {arXiv},
  primaryClass = {cs.AI},
  url          = {https://arxiv.org/abs/2503.22674},
  note         = {arXiv preprint}
}

@misc{montazeralghaem2025preference,
  title        = {Asking Clarifying Questions for Preference Elicitation With Large Language Models},
  author       = {Montazeralghaem, Ali and Tennenholtz, Guy and Boutilier, Craig and Meshi, Ofer},
  year         = {2025},
  eprint       = {2510.12015},
  archivePrefix= {arXiv},
  primaryClass = {cs.AI},
  url          = {https://arxiv.org/abs/2510.12015},
  note         = {arXiv preprint}
}

@misc{wu2023codeclarify,
  title        = {Large Language Models Should Ask Clarifying Questions to Increase Confidence in Generated Code},
  author       = {Wu, Jie JW},
  year         = {2023},
  eprint       = {2308.13507},
  archivePrefix= {arXiv},
  primaryClass = {cs.SE},
  url          = {https://arxiv.org/abs/2308.13507},
  note         = {arXiv preprint}
}

@inproceedings{aliannejadi2019qulac,
  title     = {Qulac: A Dataset for Evaluating Clarifying Questions in Conversational Search},
  author    = {Aliannejadi, Mohammad and Azzopardi, Leif and Balog, Krisztian and Sanderson, Mark},
  booktitle = {Proceedings of the 42nd International ACM SIGIR Conference on Research and Development in Information Retrieval},
  year      = {2019},
  pages     = {285--294},
  address   = {Paris, France},
  publisher = {ACM},
  doi       = {10.1145/3331184.3331226}
}

@inproceedings{zamani2020mimics,
  title     = {MIMICS: A Large-Scale Data Collection for Search Clarification},
  author    = {Zamani, Hamed and Trippas, Johanne R. and Dalton, Jeff and Radlinski, Filip},
  booktitle = {Proceedings of the 29th ACM International Conference on Information and Knowledge Management},
  year      = {2020},
  pages     = {3189--3198},
  address   = {Galway, Ireland},
  publisher = {ACM},
  doi       = {10.1145/3340531.3412772}
}

@inproceedings{rao2018evpi,
  title     = {Learning to Ask Good Questions: Ranking Clarification Questions Using Neural Expected Value of Perfect Information},
  author    = {Rao, Sudha and Daum{\'e} III, Hal},
  booktitle = {Proceedings of the 56th Annual Meeting of the Association for Computational Linguistics},
  year      = {2018},
  pages     = {3669--3680},
  address   = {Melbourne, Australia},
  publisher = {Association for Computational Linguistics},
  doi       = {10.18653/v1/P18-1340},
  url       = {https://aclanthology.org/P18-1340/}
}

@inproceedings{zhang2022multiinterest,
  title     = {Multiple Choice Questions Based Multi-Interest Policy Learning for Conversational Recommendation},
  author    = {Zhang, Yiming and Wu, Lingfei and Shen, Qi and Pang, Yitong and Wei, Zhihua and Xu, Fangli and Long, Bo and Pei, Jian},
  booktitle = {Proceedings of the ACM Web Conference 2022},
  year      = {2022},
  pages     = {2153--2162},
  address   = {Lyon, France},
  publisher = {ACM},
  doi       = {10.1145/3485447.3512088},
  url       = {https://doi.org/10.1145/3485447.3512088}
}

@inproceedings{white2021open,
  title     = {Open-domain clarification question generation without question examples},
  author    = {White, Julia and Poesia, Gabriel and Hawkins, Robert and Sadigh, Dorsa and Goodman, Noah},
  booktitle = {Proceedings of the 2021 Conference on Empirical Methods in Natural Language Processing},
  year      = {2021},
  month     = nov,
  pages     = {563--570},
  address   = {Online and Punta Cana, Dominican Republic},
  publisher = {Association for Computational Linguistics},
  doi       = {10.18653/v1/2021.emnlp-main.44},
  url       = {https://aclanthology.org/2021.emnlp-main.44/}
}

@inproceedings{saeidi2018sharc,
  title     = {Interpretation of Natural Language Rules in Conversational Machine Reading},
  author    = {Saeidi, Marzieh and Bartolo, Max and Lewis, Patrick and Singh, Sameer and Rockt{\"a}schel, Tim and Sheldon, Mike and Bouchard, Guillaume and Riedel, Sebastian},
  booktitle = {Proceedings of the 2018 Conference on Empirical Methods in Natural Language Processing},
  year      = {2018},
  pages     = {2087--2097},
  address   = {Brussels, Belgium},
  publisher = {Association for Computational Linguistics},
  doi       = {10.18653/v1/D18-1233},
  url       = {https://aclanthology.org/D18-1233/}
}

@inproceedings{min2020ambigqa,
  title     = {AmbigQA: Answering Ambiguous Open-domain Questions},
  author    = {Min, Sewon and Michael, Julian and Hajishirzi, Hannaneh and Zettlemoyer, Luke},
  booktitle = {Proceedings of the 2020 Conference on Empirical Methods in Natural Language Processing},
  year      = {2020},
  pages     = {5239--5251},
  address   = {Online},
  publisher = {Association for Computational Linguistics},
  doi       = {10.18653/v1/2020.emnlp-main.466},
  url       = {https://aclanthology.org/2020.emnlp-main.466/}
}

@misc{chi2024clarinet,
  title        = {CLARINET: Augmenting Language Models to Ask Clarification Questions for Retrieval},
  author       = {Chi, Yizhou and Lin, Jessy and Lin, Kevin and Klein, Dan},
  year         = {2024},
  eprint       = {2405.15784},
  archivePrefix= {arXiv},
  primaryClass = {cs.CL},
  url          = {https://arxiv.org/abs/2405.15784},
  note         = {arXiv preprint}
}

@misc{kuhn2023semanticuncertainty,
  title        = {Semantic Uncertainty: Linguistic Invariances for Uncertainty Estimation in Natural Language Generation},
  author       = {Kuhn, Lorenz and Gal, Yarin and Farquhar, Sebastian},
  year         = {2023},
  eprint       = {2302.09664},
  archivePrefix= {arXiv},
  primaryClass = {cs.CL},
  url          = {https://arxiv.org/abs/2302.09664},
  note         = {arXiv preprint}
}

@article{farquhar2024semanticentropy,
  title   = {Detecting hallucinations in large language models using semantic entropy},
  author  = {Farquhar, Sebastian and Kossen, Jannik and Kuhn, Lorenz and Gal, Yarin},
  journal = {Nature},
  volume  = {630},
  number  = {8017},
  pages   = {625--630},
  year    = {2024},
  doi     = {10.1038/s41586-024-07421-0},
  url     = {https://www.nature.com/articles/s41586-024-07421-0}
}

@misc{kossen2024seps,
  title        = {Semantic Entropy Probes: Robust and Cheap Hallucination Detection in LLMs},
  author       = {Kossen, Jannik and Han, Jiatong and Razzak, Muhammed and Schut, Lisa and Malik, Shreshth and Gal, Yarin},
  year         = {2024},
  eprint       = {2406.15927},
  archivePrefix= {arXiv},
  primaryClass = {cs.CL},
  url          = {https://arxiv.org/abs/2406.15927},
  note         = {arXiv preprint}
}

@inproceedings{joo2025cleanse,
  title     = {Cleanse: Uncertainty Estimation Approach Using Clustering-based Semantic Consistency in LLMs},
  author    = {Joo, Minsuh and Cho, Hyunsoo},
  booktitle = {Proceedings of the Fourth Workshop on Generation, Evaluation and Metrics (GEM\textsuperscript{2})},
  year      = {2025},
  pages     = {291--301},
  publisher = {Association for Computational Linguistics},
  address   = {Online},
  url       = {https://aclanthology.org/2025.gem-1.25/}
}

@inproceedings{aliannejadi2019asking,
  title     = {Asking Clarifying Questions in Open-Domain Information-Seeking Conversations},
  author    = {Aliannejadi, Mohammad and Zamani, Hamed and Crestani, Fabio and Croft, W. Bruce},
  booktitle = {Proceedings of the 42nd International ACM SIGIR Conference on Research and Development in Information Retrieval},
  year      = {2019},
  pages     = {475--484},
  publisher = {Association for Computing Machinery},
  address   = {New York, NY, USA},
  doi       = {10.1145/3331184.3331265},
  url       = {https://doi.org/10.1145/3331184.3331265}
}

@misc{aliannejadi2020convai3clariq,
  title        = {ConvAI3: Generating Clarifying Questions for Open-Domain Dialogue Systems (ClariQ)},
  author       = {Aliannejadi, Mohammad and Kiseleva, Julia and Chuklin, Aleksandr and Dalton, Jeff and Burtsev, Mikhail},
  year         = {2020},
  eprint       = {2009.11352},
  archivePrefix= {arXiv},
  primaryClass = {cs.CL},
  url          = {https://arxiv.org/abs/2009.11352},
  note         = {arXiv preprint / shared task overview}
}

@inproceedings{sekulic2021facetdriven,
  title     = {Towards Facet-Driven Generation of Clarifying Questions for Conversational Search},
  author    = {Sekuli{\'c}, Ivan and Aliannejadi, Mohammad and Crestani, Fabio},
  booktitle = {Proceedings of the 2021 ACM SIGIR International Conference on Theory of Information Retrieval},
  year      = {2021},
  pages     = {167--175},
  publisher = {Association for Computing Machinery},
  address   = {New York, NY, USA},
  doi       = {10.1145/3471158.3472257},
  url       = {https://doi.org/10.1145/3471158.3472257}
}

@inproceedings{majumder2021global,
  title     = {Ask What’s Missing and What’s Useful: Improving Clarification Question Generation Using Global Knowledge},
  author    = {Majumder, Bodhisattwa Prasad and Rao, Sudha and Galley, Michel and McAuley, Julian},
  booktitle = {Proceedings of the 2021 Conference of the North American Chapter of the Association for Computational Linguistics: Human Language Technologies},
  year      = {2021},
  pages     = {4300--4312},
  address   = {Online},
  publisher = {Association for Computational Linguistics},
  url       = {https://aclanthology.org/2021.naacl-main.340/}
}

@inproceedings{lee-etal-2023-asking,
  author    = {Dongryeol Lee and Segwang Kim and Minwoo Lee and Hwanhee Lee and Joonsuk Park and Sang-Woo Lee and Kyomin Jung},
  title     = {Asking Clarification Questions to Handle Ambiguity in Open-Domain QA},
  booktitle = {Findings of the Association for Computational Linguistics: EMNLP 2023},
  year      = {2023},
  pages     = {11526--11544},
  publisher = {Association for Computational Linguistics},
  address   = {Singapore},
  doi       = {10.18653/v1/2023.findings-emnlp.772},
  url       = {https://aclanthology.org/2023.findings-emnlp.772/}
}

@inproceedings{yuan2024multimodal,
  title     = {Asking Multimodal Clarifying Questions in Mixed-Initiative Conversational Search},
  author    = {Yuan, Yifei and Siro, Clemencia and Aliannejadi, Mohammad and de Rijke, Maarten and Lam, Wai},
  booktitle = {Proceedings of the ACM Web Conference 2024},
  year      = {2024},
  pages     = {1474--1485},
  publisher = {Association for Computing Machinery},
  address   = {New York, NY, USA},
  doi       = {10.1145/3589334.3645483},
  url       = {https://doi.org/10.1145/3589334.3645483}
}

@inproceedings{zhao2024intentaware,
  title     = {Generating Intent-aware Clarifying Questions in Conversational Information Retrieval Systems},
  author    = {Zhao, Ziliang and Dou, Zhicheng and Zhou, Yujia},
  booktitle = {Proceedings of the 33rd ACM International Conference on Information and Knowledge Management},
  year      = {2024},
  pages     = {3384--3394},
  publisher = {Association for Computing Machinery},
  address   = {New York, NY, USA},
  doi       = {10.1145/3627673.3679851},
  url       = {https://doi.org/10.1145/3627673.3679851}
}

@inproceedings{zhang2024clamber,
  title     = {CLAMBER: A Benchmark of Identifying and Clarifying Ambiguous Information Needs in Large Language Models},
  author    = {Zhang, Tong and Mao, Jiali and Yao, Shunyu and Wang, Rui and Cao, Yixin},
  booktitle = {Proceedings of the 62nd Annual Meeting of the Association for Computational Linguistics},
  year      = {2024},
  pages     = {10718--10735},
  address   = {Bangkok, Thailand},
  publisher = {Association for Computational Linguistics},
  url       = {https://aclanthology.org/2024.acl-long.578/}
}

@inproceedings{qian2024implicit,
  title     = {Tell Me More! Towards Implicit User Intention Understanding in Agent Interaction},
  author    = {Qian, Cheng and Liu, Yuhan and Lan, Zhenzhong and Liu, Yixuan and Zhang, Jing and Huang, Minlie},
  booktitle = {Proceedings of the 62nd Annual Meeting of the Association for Computational Linguistics},
  year      = {2024},
  pages     = {1114--1139},
  address   = {Bangkok, Thailand},
  publisher = {Association for Computational Linguistics},
  url       = {https://aclanthology.org/2024.acl-long.61/}
}

@misc{zhang2024futureturns,
  title        = {Modeling Future Conversation Turns to Teach Large Language Models to Ask Clarifying Questions},
  author       = {Zhang, Michael J. Q. and Knox, W. Bradley and Choi, Eunsol},
  year         = {2024},
  eprint       = {2410.13788},
  archivePrefix= {arXiv},
  primaryClass = {cs.CL},
  url          = {https://arxiv.org/abs/2410.13788},
  note         = {arXiv preprint; submitted to ICLR 2025}
}

@inproceedings{zhao2025clarilm,
  title     = {ClariLM: Enhancing Open-domain Clarification Ability for Large Language Models},
  author    = {Zhao, Ziliang and Chen, Haonan and Song, Shiren and Xie, Jian and Dou, Zhicheng},
  booktitle = {Proceedings of the 34th ACM International Conference on Information and Knowledge Management},
  year      = {2025},
  pages     = {4401--4411},
  publisher = {Association for Computing Machinery},
  address   = {New York, NY, USA},
  doi       = {10.1145/3746252.3761068},
  url       = {https://doi.org/10.1145/3746252.3761068}
}

@inproceedings{ma2025ambigchat,
  title     = {AmbigChat: Interactive Hierarchical Clarification for Ambiguous Open-Domain Question Answering},
  author    = {Ma, Jiaju and Shi, Lei and Robertsen, Kenneth and Chi, Peggy},
  booktitle = {Proceedings of the 38th Annual ACM Symposium on User Interface Software and Technology},
  year      = {2025},
  pages     = {141:1--141:18},
  publisher = {Association for Computing Machinery},
  address   = {New York, NY, USA},
  doi       = {10.1145/3746059.3747686},
  url       = {https://doi.org/10.1145/3746059.3747686}
}

@inproceedings{ouyang2022training,
  title     = {Training Language Models to Follow Instructions with Human Feedback},
  author    = {Ouyang, Long and Wu, Jeffrey and Jiang, Xu and Almeida, Diogo and Wainwright, Carroll L. and Mishkin, Pamela and Zhang, Chong and Agarwal, Sandhini and Slama, Katarina and Ray, Alex and Schulman, John and Hilton, Jacob and Kelton, Fraser and Miller, Luke and Simens, Maddie and Askell, Amanda and Welinder, Peter and Christiano, Paul F. and Leike, Jan and Lowe, Ryan},
  booktitle = {Advances in Neural Information Processing Systems},
  year      = {2022},
  volume    = {35},
  pages     = {27730--27744},
  publisher = {Curran Associates, Inc.},
  address   = {Red Hook, NY, USA},
  doi       = {10.5555/3600270.3602281},
  url       = {https://dl.acm.org/doi/10.5555/3600270.3602281}
}

@article{shannon1948entropy,
  title   = {A Mathematical Theory of Communication},
  author  = {Shannon, Claude E.},
  journal = {Bell System Technical Journal},
  year    = {1948},
  volume  = {27},
  number  = {3},
  pages   = {379--423, 623--656}
}

@inproceedings{rafailov2023dpo,
  title     = {Direct Preference Optimization: Your Language Model Is Secretly a Reward Model},
  author    = {Rafailov, Rafael and Sharma, Archit and Mitchell, Eric and Ermon, Stefano and Finn, Chelsea and Levine, Sergey},
  booktitle = {Advances in Neural Information Processing Systems},
  year      = {2023},
  volume    = {36},
  publisher = {Curran Associates, Inc.},
  address   = {Red Hook, NY, USA},
  url       = {https://arxiv.org/abs/2305.18290}
}

@misc{shao2024deepseekmath,
  title        = {DeepSeekMath: Pushing the Limits of Mathematical Reasoning in Open Language Models},
  author       = {Shao, Zhihong and Wang, Peiyi and Zhu, Qihao and Xu, Runxin and Song, Junxiao and Bi, Xiao and Zhang, Haowei and Zhang, Mingchuan and Li, Y.~K. and Wu, Y. and Guo, Daya},
  year         = {2024},
  eprint       = {2402.03300},
  archivePrefix= {arXiv},
  primaryClass = {cs.CL},
  url          = {https://arxiv.org/abs/2402.03300},
  note         = {arXiv preprint}
}

@inproceedings{wang2023selfconsistency,
  title     = {Self-Consistency Improves Chain-of-Thought Reasoning in Language Models},
  author    = {Wang, Xuezhi and Wei, Jason and Schuurmans, Dale and Bosma, Maarten and Le, Quoc V. and Chi, Ed H. and Narang, Sharan and Chowdhery, Aakanksha and Zhou, Denny},
  booktitle = {Proceedings of the International Conference on Learning Representations},
  year      = {2023},
  url       = {https://openreview.net/forum?id=1PL1NIMMrw}
}

@misc{kadavath2022language,
  title        = {Language Models (Mostly) Know What They Know},
  author       = {Kadavath, Saurabh and Arora, Aman and Schulman, John and Henighan, Tom and Steinhardt, Jacob and Kaplan, Jared and Dhariwal, Prafulla and Amodei, Dario},
  year         = {2022},
  eprint       = {2207.05221},
  archivePrefix= {arXiv},
  primaryClass = {cs.CL},
  url          = {https://arxiv.org/abs/2207.05221},
  note         = {arXiv preprint}
}

@misc{rothe2017question_as_program,
  title        = {Question Asking as Program Generation},
  author       = {Rothe, Anselm and Lake, Brenden M. and Gureckis, Todd M.},
  year         = {2017},
  eprint       = {1711.06351},
  archivePrefix= {arXiv},
  primaryClass = {cs.CL},
  url          = {https://arxiv.org/abs/1711.06351},
  note         = {arXiv preprint}
}

@inproceedings{liu2024mining,
  title     = {Mining Exploratory Queries for Conversational Search},
  author    = {Liu, Wenhan and Zhao, Ziliang and Zhu, Yutao and Dou, Zhicheng},
  booktitle = {Proceedings of The Web Conference 2024},
  year      = {2024},
  pages     = {1386--1394},
  publisher = {Association for Computing Machinery},
  address   = {New York, NY, USA},
  doi       = {10.1145/3589334.3645424},
  url       = {https://doi.org/10.1145/3589334.3645424}
}

@misc{gu2024survey_llm_as_judge,
  title        = {A Survey on LLM‐as‐a‐Judge},
  author       = {Gu, Jiawei and Jiang, Xuhui and Shi, Zhichao and Tan, Hexiang and Zhai, Xuehao and Xu, Chengjin and Li, Wei and Shen, Yinghan and Ma, Shengjie and Liu, Honghao and Wang, Saizhuo and Zhang, Kun and Wang, Yuanzhuo and Ni, Lionel and Guo, Jian and Gao, Wen},
  year         = {2024},
  eprint       = {2411.15594},
  archivePrefix= {arXiv},
  primaryClass = {cs.CL},
  url          = {https://arxiv.org/abs/2411.15594},
  note         = {arXiv preprint}
}

@misc{dou2025pretrained,
  title        = {Pre-Trained Policy Discriminators are General Reward Models},
  author       = {Dou, Shihan and Liu, Shichun and Yang, Yuming and Zou, Yicheng and Zhou, Yunhua and Xing, Shuhao and Huang, Chenhao and Ge, Qiming and Song, Demin and Lv, Haijun and Gao, Songyang and Lv, Chengqi and Zhou, Enyu and Guo, Honglin and Xi, Zhiheng and Zhang, Wenwei and Guo, Qipeng and Zhang, Qi and Qiu, Xipeng and Huang, Xuanjing and Gui, Tao and Chen, Kai},
  year         = {2025},
  eprint       = {2507.05197},
  archivePrefix= {arXiv},
  primaryClass = {cs.LG},
  url          = {https://arxiv.org/abs/2507.05197},
  note         = {arXiv preprint}
}

\end{document}